\documentclass[journal]{IEEEtai}
\IEEEoverridecommandlockouts
\usepackage[colorlinks,urlcolor=blue,linkcolor=blue,citecolor=blue]{hyperref}
\usepackage{color,array}
\usepackage{graphicx}

\usepackage{mathtools}
\usepackage{algorithm}
\usepackage{algorithmic}
\makeatother
\usepackage[utf8]{inputenc}
\usepackage[table]{xcolor}
\usepackage{tabularx,booktabs}

\newcolumntype{C}{>{\centering\arraybackslash}X} 
\usepackage{cite}

\usepackage{amsmath}
\usepackage{amsmath,amssymb,amsfonts}
\usepackage{algorithmic}
\usepackage{graphicx}
\usepackage{textcomp}
\usepackage{xcolor}
\usepackage{hyperref}
\usepackage{caption}
\usepackage{pifont}
\usepackage{xcolor}
\usepackage{rotating}
\usepackage{rotfloat}
\usepackage{algorithm}
\usepackage{algorithmic}
\usepackage{gensymb}
\usepackage{textcomp}
\usepackage{placeins}
\usepackage{balance}
\usepackage{subfigure}
\usepackage{booktabs}
\usepackage{tabularx}
\usepackage{array}
\usepackage{pbox}
\usepackage{longtable}
\usepackage{booktabs}
\def\BibTeX{{\rm B\kern-.05em{\sc i\kern-.025em b}\kern-.08em
    T\kern-.1667em\lower.7ex\hbox{E}\kern-.125emX}}
\begin{document}

\author{\IEEEauthorblockN{Praneeth Nemani, Satyanarayana Vollala}\\
\IEEEauthorblockA{\textit{Dept. of Computer Science and Engineering} \\
\textit{IIIT Naya Raipur}\\
Raipur, Chhattisgarh, India \\
praneeth19100, satya@iiitnr.edu.in}}

\title{A Cognitive Study on Semantic Similarity Analysis of Large Corpora: A Transformer-based Approach} 

\IEEEoverridecommandlockouts
\IEEEpubid{\makebox[\columnwidth]{978-1-6654-7350-7/22/\$31.00~\copyright2022 IEEE \hfill}
\hspace{\columnsep}\makebox[\columnwidth]{ }}

\maketitle
\IEEEpubidadjcol
\begin{abstract}
Semantic similarity analysis and modeling is a fundamentally acclaimed task in many pioneering applications of natural language processing today. Owing to the sensation of sequential pattern recognition, many neural networks like RNNs and LSTMs have achieved satisfactory results in semantic similarity modeling. However, these solutions are considered inefficient due to their inability to process information in a non-sequential manner, thus leading to the improper extraction of context. Transformers function as the state-of-the-art architecture due to their advantages like non-sequential data processing and self-attention. In this paper, we perform semantic similarity analysis and modeling on the U.S Patent Phrase to Phrase Matching Dataset using both traditional and transformer-based techniques. We experiment upon four different variants of the Decoding Enhanced BERT - DeBERTa and enhance its performance by performing K-Fold Cross-Validation. The experimental results demonstrate our methodology's enhanced performance compared to traditional techniques, with an average Pearson correlation score of 0.79. 
\end{abstract}

\begin{IEEEkeywords}
Semantic Similarity, K-Fold Cross Validation, Pearson Correlation, Transformers
\end{IEEEkeywords}

\section{Introduction}
\begin{figure}[htbp]
    \centering
    \includegraphics[width = \linewidth]{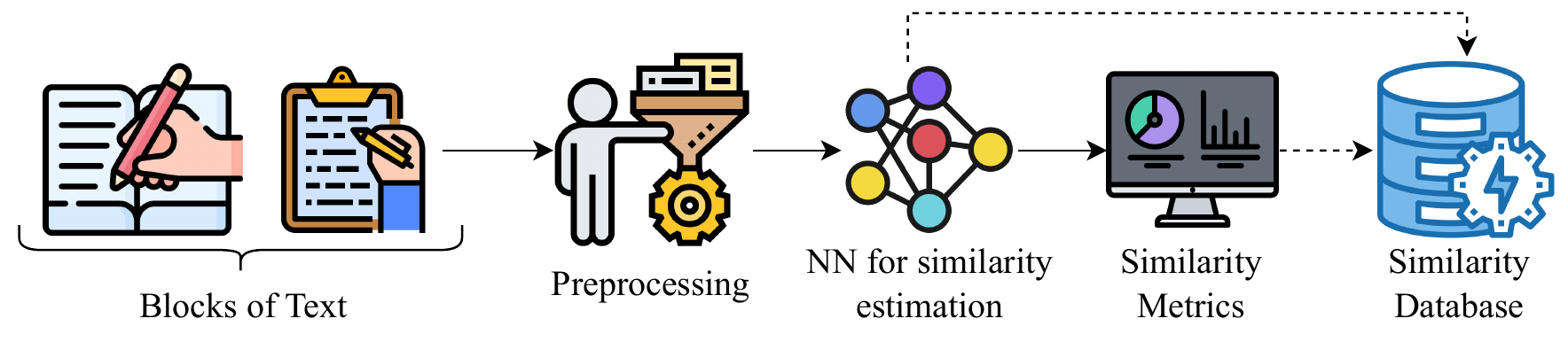}
    \caption{Conceptual Overview of Semantic Similarity Analysis}
    \label{fig:CO}
\end{figure}

Semantic similarity is defined as the association between two blocks of text, including sentences, words, and documents. It plays a fundamentally acclaimed role in most of the NLP tasks performed by researchers worldwide today. The dynamic and versatile nature of human language makes it difficult to standardize the process of semantic similarity \cite{chandrasekaran2021evolution}. As time evolves, finding new semantic analysis techniques is deemed essential due to the exponential rise of textual data generation. The conceptual overview of semantic similarity analysis is depicted in Fig. \ref{fig:CO}. 

As mentioned above, semantic similarity analysis is pivotal in various applications like information recovery, text summarization, speech enhancement, and automatic dialogue generation. In the initial methodologies proposed by researchers worldwide, semantic similarity was calculated based on the number of similar words in two blocks of text. However, this yielded inaccurate results as there were instances where two blocks of text had many similar words but conveyed different meanings. For example, the sentences {\it Tom and Harry played Badminton and Cricket}, and {\it Tom played Badminton and Harry played Cricket} are lexically similar. However, the context of these two sentences was not the same. Conversely, the sentences {\it Jenny knows many languages} and {\it Jenny is a polyglot}, though lexically dissimilar, convey the same meaning. Similarity analysis based on lexical similarity methods is easy to implement, but they fail in sentences that are lexically dissimilar and semantically similar.

Today's pioneering ML algorithms and their applications use the method of vectorization for the process of feature extraction. The concept of vectorization in NLP is that each word/phrase in a dataset is represented as a vector or an array of numbers, making feature extraction easier owing to today's computers' computational efficiency. Some techniques that use this vectorization concept include Bag of Words \cite{zhang2010understanding} and TF-IDF \cite{qaiser2018text}. However, the major limitation of these solutions is that they do not consider the broader context of the text block for computing semantic similarity. The broader context between two blocks of text is inversely proportional to their semantic distance. 

Recurrent Neural Networks and Long Short-Term Memory networks, abbreviated as RNNs and LSTMs, respectively, have been considered effective techniques for learning dependencies between blocks of text. While RNNs depend on the recent previous blocks, LSTMs depend on the broader context of the text. However, they are considered inferior to transformers due to their limitation of sequential data processing. The significant advantages of transformers like the training of large corpus, non-sequential data processing, self-attention, and techniques like positional embeddings to replace recurrent have popularized them for performing modern-day NLP tasks. This paper provides an insight into how modern-day transformers can be used for the task of semantic similarity analysis. The major contributions of our paper can be listed below:

\begin{itemize}
    \item We present a comprehensive study on the different methodologies used for the process of semantic similarity analysis. 
    \item We also perform the state-of-the-art preprocessing techniques and exploratory data analysis on the U.S Patent Phrase-to-Phrase matching dataset. 
    \item Exclusive experimentation and analysis of one traditional and four transformer-based techniques is performed to extract context and perform semantic similarity analysis. 
\end{itemize}

\section{Related Research Overview}
Research on Semantic Analysis has been a topic of interest since the 20th century. Many solutions have been proposed with their implementation on many benchmark datasets. This section presents a comprehensive survey of datasets and different methodologies used for Semantic Similarity Analysis. Table \ref{Datasets} gives an overview of the widely-used datasets for semantic similarity. 

\begin{table}[htbp]
\caption{Popular Datasets for semantic similarity}
\resizebox{\columnwidth}{!}{%
\begin{tabular}{|c|c|c|c|}
\hline
\textbf{Author}&
\textbf{Dataset}& 
\textbf{Word/Sentence pairs}&
\textbf{Similarity Score Range}\\ \hline
Rubenstein et al. \cite{rubenstein1965contextual} & 
R\& G  & 
65&
0-4 \\ 
\hline

Miller et al. \cite{miller1991contextual} &
M\&C &
30 &
0-4 \\ 
\hline

Finkelstein et al. \cite{finkelstein2001placing} &
WS353 &
353 &
0-10 \\ 
\hline

Agirre et al. \cite{agirre2015semeval} &
STS2015 &
3000 &
0-5 \\ 
\hline

Marelli et al. \cite{marelli2014sick} &
SICK &
10000 &
1-5 \\
\hline

USTPO &
Patent Phrase Matching &
33000 &
0-1 \\ 
\hline
\end{tabular}
}
\label{Datasets}
\end{table}

Benajiba et al. \cite{benajiba2019siamese} proposed a solution involving a Siamese LSTM regression model that is used to predict the similarity of the SQL template of two questions. Using the proposed methodology, the authors defined a metric called the SQL structure distance used to estimate the similarity. To reduce the computational cost of the solution, the authors have clustered the training set samples with the one-hot lexical representation of the questions. Li et al. \cite{9057451} proposed another solution involving semantic similarity in biomedical sentences using an SNN approach. The methodology involved the integration of an interactive self-attention (ISA) mechanism and an SNN. The proposed solution was validated on three standard biomedical datasets with an average Pearson score of 0.65. Pontes et al. \cite{DBLP:journals/corr/abs-1810-10641} proposed a Siamese CNN + LSTM model in which the CNN extracts the local context while the LSTM extracts the global context. The proposed methodology was evaluated on the SICK dataset with different combinations of local context and global context. 

Quan et al. \cite{8642425} proposed a framework combining the capability of word embeddings and attention weight mechanism by integrating them into a unified network known as the Attention Constituency Vector Tree (ACVT). The proposed solution was validated on 19 benchmark datasets which include STS’12-STS’15 with a Pearson score of 0.75. Shancheng et al. \cite{8332874} proposed a double sequential network consisting of identical LSTM layers that simultaneously train two sequences of sentences. The outputs of both the layers were passed through the dense layer and compressed to obtain the semantic similarity. The proposed solution addressed the problem of Chinese characteristics and was compared with the Baidu Semantic Text Similarity model and achieved higher accuracy. Yang et al. \cite{9094995} proposed a methodology that involved an extensive semantic network known as Probase. From the current weights and parameters of probase, the semantic similarity was performed on the M\&G, WS353-Sim, and R\&G datasets. 

In recent years, Generative Adversarial Networks (GANs) have gained tremendous popularity in artificial data generation for various tasks, including image sample generation with limited data \cite{9719383} and text generation. In this view, Liang et al. \cite{9509434} addressed the generation and identification of similar sentences using a GAN-based approach. The authors proposed a  syntactic and semantic long short-term memory (SSLSTM) algorithm for evaluating semantic similarity. Three variations of the sentence similarity generative adversarial network (SSGAN) algorithm were proposed for generating sentences. 
The state-of-the-art solutions for tasks in natural language processing involve the usage of transformers. Precisely, transformers in NLP are used to solve NLP tasks involving the dependency of long sequences. In this context, Li et al. \cite{9313452} introduced a hybrid Cross2self attention, Bi-RNN - BERT model to computer semantic similarity in biomedical data. The methodology was validated on the OHNLP2018 baselines with an increase of 0.6\% in the Pearson coefficient. Another approach involving the usage of BERT for semantic similarity of outlook emails was proposed by Sanjeev et al \cite{9212979}. Some of the standard approaches in NLP for semantic analysis include Word2Vec, proposed by Google in 2013 and the Glove model. The related research overview could be summarized in Table \ref{related}.

\begin{table}[htbp]
\caption{Overview of the existing solutions}
\begin{center}
\resizebox{\columnwidth}{!}{%
\begin{tabular}{|c|p{42mm}|p{32mm}|}
\hline
\textbf{Author}&
\centering \textbf{Methodology}& 
\centering\arraybackslash \textbf{Dataset used}\\ \hline

Benajiba et al. \cite{benajiba2019siamese} &
\centering Siamese LSTM Regression &
\centering\arraybackslash WikiSQL \\ \hline

Li et al. \cite{9057451} &
\centering ISA + Siamese NNs &
\centering\arraybackslash DBMI , CDD-ful, CDD-ref \\ \hline

Pontes et al. \cite{DBLP:journals/corr/abs-1810-10641} &
\centering Siamese CNN + LSTM &
\centering\arraybackslash SICK dataset \\ \hline

Quan et al. \cite{8642425} &
\centering Attention Constituency Vector Tree (ACVT) &
\centering\arraybackslash STS’12-STS’15 \\ \hline

Shancheng et al. \cite{8332874} &
\centering Double Seq. NN + LSTM &
\centering\arraybackslash Chinese semantic similarity dataset \\ \hline

Yang et al. \cite{9094995} &
\centering Probase &
\centering\arraybackslash M\&G, WS353-Sim, and R\&G \\ \hline

Liang et al. \cite{9509434} &
\centering SSLSTM + SSGAN & 
\centering\arraybackslash SemEval and Quora \\ \hline

Li et al. \cite{9313452} &
\centering Cross2self, BERT & 
\centering\arraybackslash Biomedical Data \\ \hline

Sanjeev et al \cite{9212979} &
\centering BERT &
\centering\arraybackslash Outlook Emails \\ \hline

\textbf{Our Work} &
\centering \textbf{DeBERTa + K-Fold Stratified Cross Validation} & 
\centering\arraybackslash \textbf{U.S Patent Phrase to Phrase Matching Dataset} \\ \hline

\end{tabular}
}
\end{center}
\label{related}
\end{table}
\vspace{-16px}

\section{Methodology}
This section deals with the different techniques used to perform the task semantic similarity analysis. In this work, we compare and analyze the performance of five different techniques used for semantic similarity analysis. These include Levenshtein Metric similarity and four different variants of the DeBERTa model. This section deals with the architecture of each model and how it can be fine-tuned for our dataset to perform the required task. 
\vspace{-12px}

\subsection{Levenshtein Metric}
In Natural Language Processing, the Levenshtein distance between two words is defined as the number of single-character edits required to convert one word from other \cite{haldar2011levenshtein}. It is a string metric used to understand the disparity between two different sequences. Edits can be defined as insertion, replacement, and deletion in this context. Some of the Levenshtein Distance applications include DNA Analysis and Plagiarism Checking. In this task of semantic similarity analysis, we experiment with the approach of the Levenshtein Distance on our dataset. 

\subsection{DeBERTa}

As mentioned in section II, there has been a remarkable rise in the usage of transformers in many NLP tasks like semantic analysis and dialogue generation. BERT (Bidirectional Encoder Representations from Transformers) has been acknowledged as a recent advancement in transformers by researchers at Google in their work \cite{vaswani2017attention}. The concept of BERT lies in the fact that when sequential data is trained in a bi-directional manner, better and deeper inference can be obtained on the data for specific tasks like language understanding. In Machine Vision, transfer learning is a widely used technique by researchers across the globe to perform various tasks rather than training a model from the onset. The idea of transfer learning is that existing deep learning models could be transformed into objective-specific models by fine-tuning the existing model. This approach has gained significance among NLP researchers worldwide, and transfer learning could now be applied to many NLP tasks. BERT employs the usage of a transformer, a mechanism that is based on attention that comprehends the contextual inference between two words in a corpus. The simplest form of BERT has an encoder and a decoder. The purpose of the encoder is to comprehend the input text, while the decoder's purpose is to deliver a prediction. 

Since 2018, there has been a rapid rise in the design and development of pre-trained language models like GPT, T5, RoBERTa, StructBERT, and DeBERTa \cite{he2021deberta}. However, in this work, we emphasize the different versions of DeBERTa and their performance in the U.S Patent to Phrase matching dataset. DeBERTa is a Decoding Enhanced BERT with disentangled attention, which functions based on introducing two novel techniques: Disentangled attention and enhanced masked decoding. The concept of disentangled attention is that each word or token in the input layer is represented by two vectors corresponding to its content and position in the corpus. This is inferred from the fact that the word's position also has significant importance in content extraction. However, though disentangled attention conveys the relative positions of words, it is deemed essential to determine the exact position of words in a corpus to avoid semantic disparity. So to achieve this, DeBERTa integrates the positional word embeddings prior to its softmax layer. Owing to its architecture, DeBERTa is considered significantly superior to its counterparts like RoBERTa \cite{liu2019roberta}. 

The input to this pipeline is the \textit{anchor phrase + seperated token + the context phrase}. DeBERTa uses a metric known as the cross-attention score to infer the semantic similarity between two blocks of text. Mathematically, the cross attention score of a block \textit{m} with respect to another block \textit{n} can be represented as shown in Eq. \ref{eq: DeBERTa} where $C_{m}$ represents the content of the word and $Pos_{m|n}$ represents the position of the word \textit{m} with respect to \textit{n}. The cross-attention score between two blocks \textit{m} and \textit{n} can be categorized into four components: {\it block value-to-block value}, {\it block value-to-index}, { \it index-to-index}, and {\it index-to-block value.} as shown in Eq. \ref{eq: DeBERTa}

\begin{equation}
\label{eq: DeBERTa}
 \begin{multlined}
    S_{m,n} = [C_{m,n}, Pos_{m|n}] \times [C_{n,m}, Pos_{n|m}]^{T} \\ 
            = C_{m}{C_{n}^{T}} + C_{m}Pos_{n|m}^{T} + Pos_{m|n}{C_{n}^{T}} + Pos_{m|n}Pos_{n|m}^{T}
    \end{multlined}
\end{equation}

However, there have been recent improvements in the composition of DeBERTa owing to ELECTRA-Style Pre-Training \cite{he2021debertav3}. The version, also known as DeBERTa-V3, has many variants, including DeBERTa-base, DeBERTa-V3-Small, DeBERTa-V3-XSmall, and mDeBERTa-V3-Base. The initial version of DeBERTa uses a mask language modeling (MLM) mechanism, which is now replaced by replaced token detection (RTD), considered to be a sample-efficient pre-training task. The variants of DeBERTa differ in their backbone parameters, vocabulary, hidden size, and layers. The architectural specifications of all the variants have been depicted in Table \ref{DeBERTaversions}. Once we input the data into the model, we now perform the task of Stratified K-Fold cross-validation. 

\begin{table}[htbp]
\centering
\caption{Specifications of the different versions of DeBERTa}
\begin{center}
\resizebox{\columnwidth}{!}{%
\centering
\begin{tabular}{|c|c|p{16mm}|c|c|}
\hline
\centering \textbf{Model}&
\textbf{Corpus}& 
\centering \textbf{Backbone Parameters(M)}&
\textbf{Hidden Size}&
\textbf{Layers}
\\ \hline
\centering DeBERTa-V3-Base &
128 & 
\centering 86 & 
768 &
12 \\ 
\hline

\centering DeBERTa-V3-Small &
128 &
\centering 44 &
768 &
6 \\
\hline

\centering DeBERTa-V3-XSmall &
128 &
\centering 	22 &
384 &
12 \\
\hline

\centering mDeBERTa-V3-Base &
250 &
\centering 86 &
768 & 
12 \\ 
\hline

\end{tabular}
}
\end{center}
\label{DeBERTaversions}
\end{table}
\vspace{-16px}

\begin{figure*}[t]
    \centering
    \includegraphics[width = \linewidth]{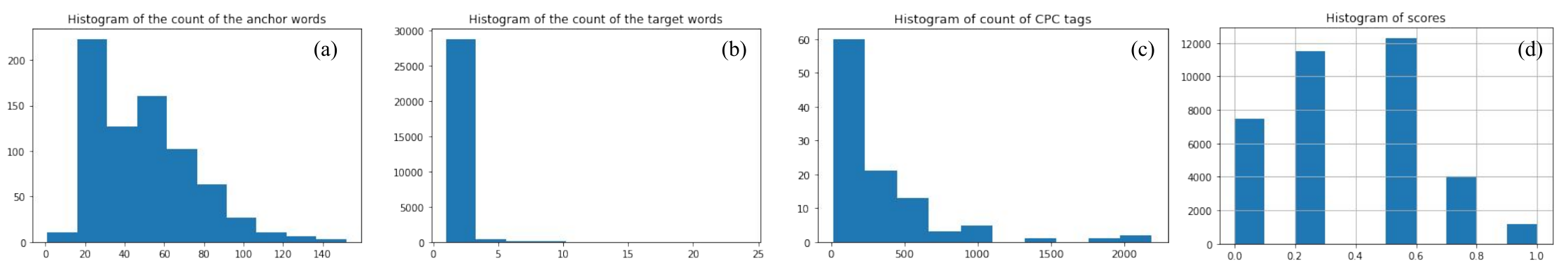}
    \caption{Distribution of Terms in (a) Anchor Phrases, (b) Target Phrases, (c) Context Tags. (d) Distribution of scores}
    \label{fig:EDA}
\end{figure*}

\begin{figure*}[t]
    \centering
    \includegraphics[width = \linewidth]{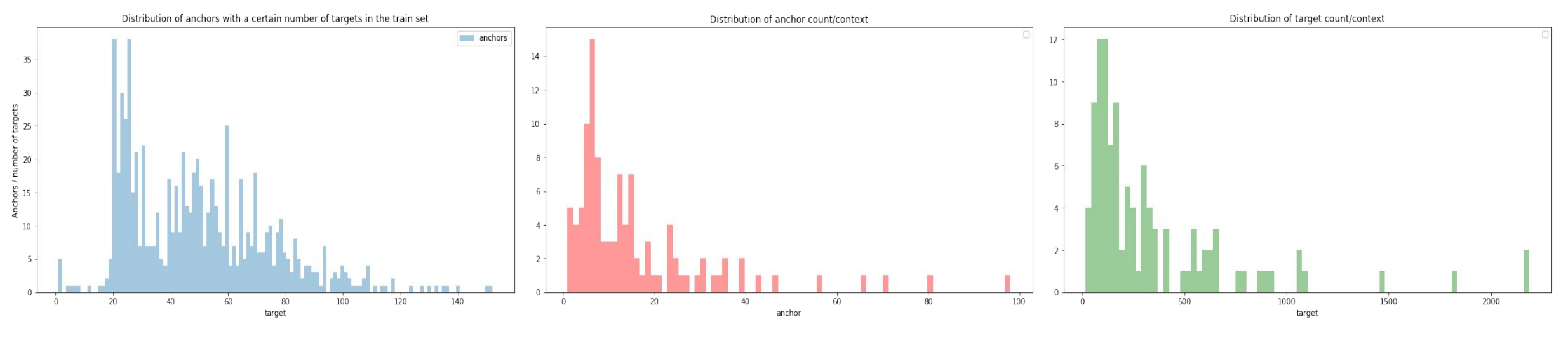}
    \caption{(a) Distribution of anchors with respect to targets, (b) Distribution of anchor count with respect to context, (c) Distribution of target count with respect to context}
    \label{fig:EDA2}
\end{figure*}

\subsection{Stratified K-Fold Cross Validation}
It is deemed essential to evaluate our model once trained on our input data. A methodological error is incorporated if the model retains the parameters of a periodic function and is experimented on the same data. The prediction scores would remain perfect on known labels, and the model's performance would still be unsatisfactory on unseen data. This condition is also known as overfitting. So to prevent overfitting, it is always deemed essential to split out a chunk of data into the test/validation set. However, there is a probability of overfitting the test/validation set due to tweaking the existing parameters until the estimator performs correctly. So to address this situation, we perform $K-Fold$ Stratified Cross-Validation \cite{8012491}. The objective of the $K-Fold$ Stratified Cross-Validation is that the data is split into $K$ folds. Training is performed on $K-1$ folds while testing is performed on the remaining fold, resulting in a higher performance of the model. Mathematically, the cross-validation estimate $CV$ can be represented in Eq. \ref{Kfold}

\begin{equation}
    CV = 1/N\sum_{i=1}^{N}L(y_{i},f^{K_{i}}(x_{i}))
    \label{Kfold}
\end{equation}
where $y_{i}$ depicts the actual score, $f^{K_{i}}(x_{i})$ depicts the prediction the on $K_{i}$th fold and $L$ is the loss function. Subsequently, we evaluate the model using the Pearson correlation coefficient. 

\section{Results and Experimentation}

\subsection{Dataset Used}
\begin{figure}[htbp]
    \centering
    \includegraphics[width = \linewidth]{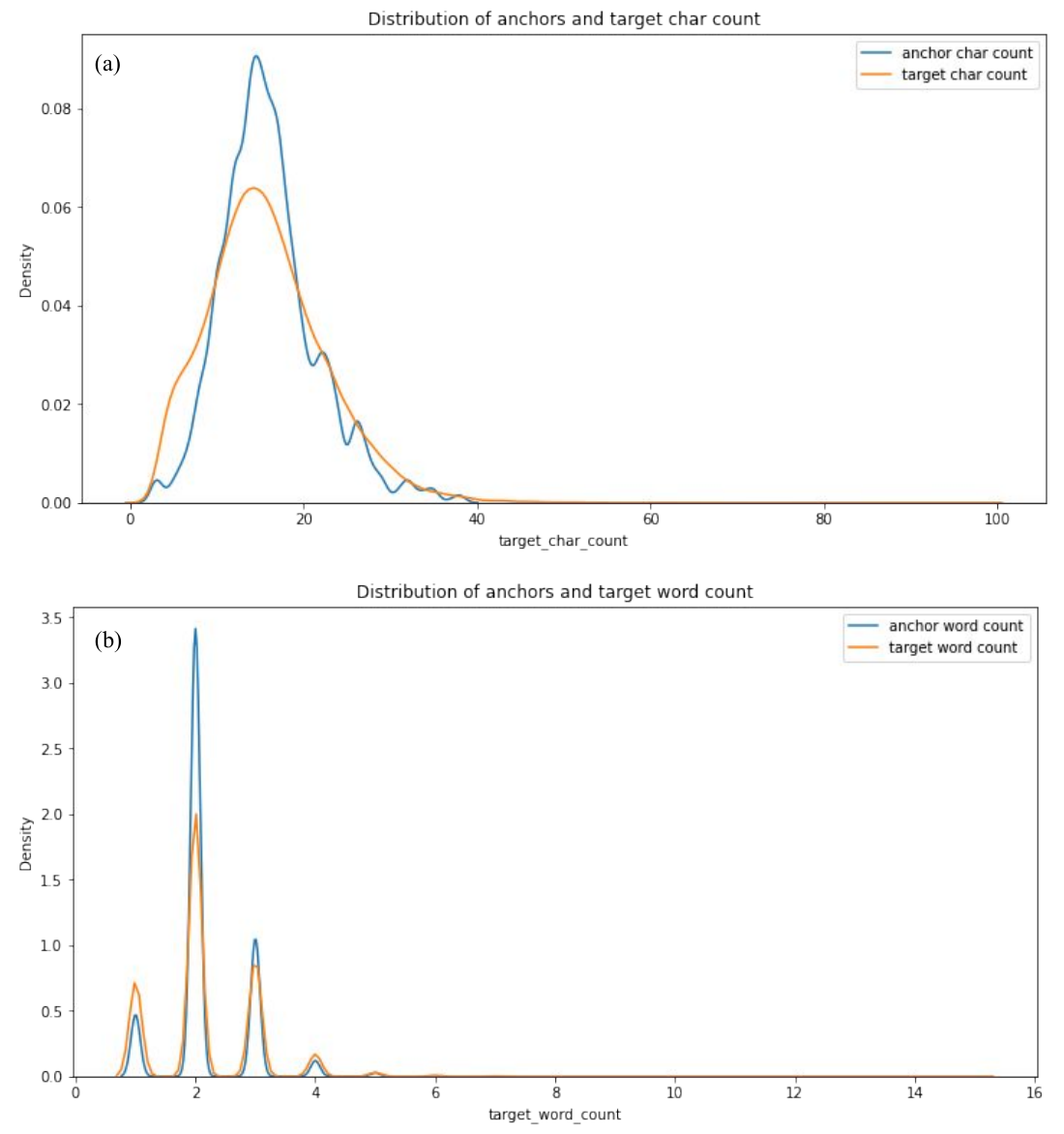}
    \caption{(a) Distribution of Anchors and Target's character count (b) word count}
    \label{fig:density}
\end{figure}

This section deals with the description and exploratory data analysis of the dataset used. We use the U.S Patent Phrase to Phrase Matching dataset to perform semantic similarity analysis in this work. The U.S Patent Phrase to Phrase Matching dataset is derived from the repositories of the U.S. Patent and Trademark Office (USPTO), and its patent archives stand as a rare blend of information volume quality, and variety. The dataset consists of 4 columns: Anchor, Target, Context, and Score. The first phrase is represented by the anchor columns, the second by the target columns, and the context column represents the subject within which the similarity is to be scored. The dataset consists of 733 unique anchor words
 and 29340 unique target words. The frequency distribution of terms of anchor, target, and context columns is depicted in Fig. \ref{fig:EDA} (a), \ref{fig:EDA} (b) and \ref{fig:EDA} (c) respectively. The distribution of the score column is represented in Fig. \ref{fig:EDA} (d). Also, we analyze the distribution of anchor phrases with respect to context and target terms.Fig. \ref{fig:EDA2} (a) shows the distribution of anchors with respect to targets, Fig. \ref{fig:EDA2} (b), the distribution of anchor count with respect to context and Fig. \ref{fig:EDA2} (c) depicts the distribution of target count with respect to context. Similarly, the character count and word count distribution of both anchor and target columns are illustrated in Fig. \ref{fig:density} (a) and \ref{fig:density} (b) respectively. 

\begin{figure*}[t]
    \centering
    \includegraphics[width = \linewidth]{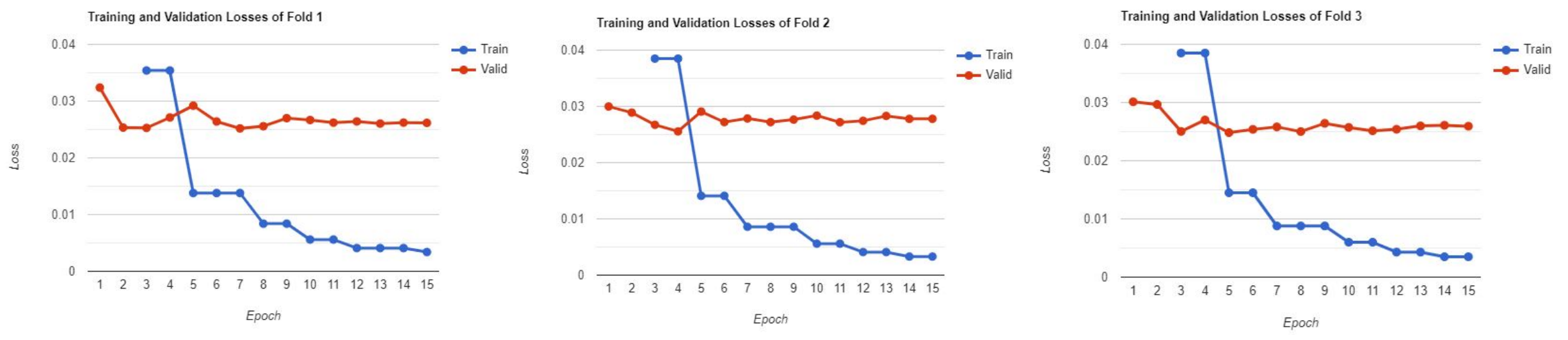}
    \caption{Training and Validation Losses of DeBERTa-small in (a) Fold 1, (b) Fold 2 and (c) Fold 3}
    \label{fig:TL}
\end{figure*}

\begin{figure*}[t]
    \centering
    \includegraphics[width = \linewidth]{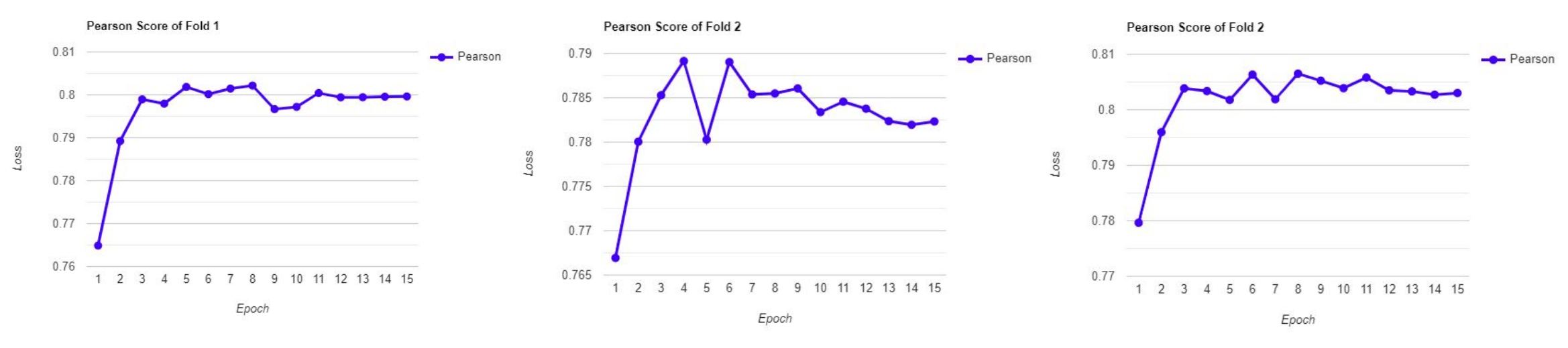}
    \caption{Variation of pearson scores of DeBERTa-small in (a) Fold 1, (b) Fold 2 and (c) Fold 3}
    \label{fig:P}
\end{figure*}

\begin{figure}[htbp]
    \centering
    \includegraphics[width = \linewidth]{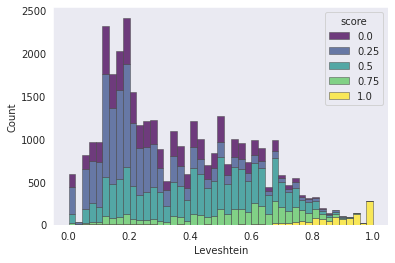}
    \caption{Distribution of Levenshtein Similarity Scores}
    \label{fig:LSS}
\end{figure}

\subsection{Results of Semantic Analysis by Levenshtein Metric}
The below Fig. \ref{fig:LSS} shows the distribution of the Levenshtein similarity score with respect to the number of anchor-target pairs. The figure shows that maximum pairs are present between 0.1 and 0.2, with the least number of pairs having the Levenshtein similarity score between 0.8 and 1.0. From this experiment, we conclude that the methodology yields us a pearson correlation score of 0.4147.

\subsection{Performance of DeBERTa-base}
In this section, we illustrate the performance of the DeBERTa-base model on the dataset. We use three metrics to evaluate the performance: training loss, validation, and Pearson correlation coefficient. The training loss measures how deep the model fits the training data and how accurate the model's predictions are on the training set. Similarly, we define the validation loss as the model's performance on the validation/test set. The final metric, also known as the Pearson correlation coefficient, gives us a linear measure of the strength of two variables. As depicted in Table \ref{DeBERTa-base}, DeBERTa-base achieved an average training loss of 0.03, validation loss of 0.026, and Pearson correlation score of 0.74. In the subsequent sections, we analyze the performance of DeBERTa-V3-Small, DeBERTa-V3-XSmall, and mDeBERTa-V3-Base.

\begin{table}[htbp]
\caption{Performance Metrics of DeBERTa-base}
\begin{center}
\resizebox{\columnwidth}{!}{%
\begin{tabular}{|c|c|c|c|}
\hline
\textbf{Fold}&
\textbf{Training Loss}& 
\textbf{Validation Loss}&
\textbf{Pearson Correlation}\\ \hline
1& 
0.030000 & 
0.024792 &
0.771547 \\ 
\hline

2& 
0.030700 & 
0.027380 &
0.760287 \\ 
\hline

3& 
0.030500 & 
0.024049 &
0.751671 \\ 
\hline

4& 
0.029900 & 
0.028247 &
0.785597 \\ 
\hline

\end{tabular}
}
\end{center}
\label{DeBERTa-base}
\end{table}

\subsection{Performance of mDeBERTa}
In this section, we emphasize the performance of multilingual DeBERTa on the training and validation sets. The number of epochs are 5, with the batch size being 128. Despite performing stratified K-Fold cross-validation with the number of folds set as 4, the model showed an inferior performance in terms of similarity prediction with very less Pearson coefficient score in all folds, as depicted in Table \ref{mDeBERTa}. This can be justified by the fact that mDeBERTa is trained upon the CC100 multilingual data, and the presence of multilingual backbone parameters led to the depreciated performance of the model. 

\begin{table}[htbp]
\caption{Performance Metrics of mDeBERTa}
\begin{center}
\resizebox{\columnwidth}{!}{%
\begin{tabular}{|c|c|c|c|}
\hline
\textbf{Fold}&
\textbf{Training Loss}& 
\textbf{Validation Loss}&
\textbf{Pearson Correlation}\\ \hline
1& 
0.273200 & 
0.276361 &
0.116614 \\ 
\hline

2& 
0.148500 & 
0.141006 &
0.154153 \\ 
\hline

3& 
0.147500 & 
0.140739 &
0.193211 \\ 
\hline

4& 
0.150100 & 
0.136254 &
0.175404 \\ 
\hline

\end{tabular}
}
\end{center}
\label{mDeBERTa}
\end{table}
\vspace{-16px}

\subsection{Performance of DeBERTa-Small}
DeBERTa-Small is an abridged version of the DeBERTa-base, keeping in view the critical parameters required for prediction. It is trained on 160GB data as its previous version with 44M backbone parameters and has a hidden size of 768 with the number of layers as 6. The model achieved a cumulative Pearson score of 0.78 after three cross-validation folds. The performance metrics of DeBERTa-Small are depicted in Table \ref{DeBERTa-Small}. From Table \ref{DeBERTa-Small}, we can infer that the best performance is illustrated by DeBERTa-Small. Also, we illustrate the training and validation losses of each fold in Fig. \ref{fig:TL} and the variation of the pearson score in Fig. \ref{fig:P}.   

\begin{table}[htbp]
\caption{Performance Metrics of DeBERTa-v3-Small}
\begin{center}
\resizebox{\columnwidth}{!}{%
\begin{tabular}{|c|c|c|c|}
\hline
\textbf{Fold}&
\textbf{Training Loss}& 
\textbf{Validation Loss}&
\textbf{Pearson Correlation}\\ \hline
1& 
0.003400 & 
0.026166 &
0.799629 \\
\hline

2& 
0.003300 &
0.027797 &
0.782329 \\ 
\hline

3& 
0.003500 &
0.025930 &
0.803020 \\
\hline

\end{tabular}
}
\end{center}
\label{DeBERTa-Small}
\end{table}

\subsection{Performance of DeBERTa-XSmall}
DeBERTa-XSmall is considered a simplified version of DeBERTa-Small with only 22M backbone parameters which is half in number compared to its counterpart. This model achieved a cumulative Pearson score of 0.765 after four cross-validation folds. However, fewer backbone parameters and hidden size justify its lesser performance than DeBERTa-Small. The performance metrics of DeBERTa-XSmall are depicted in Table \ref{DeBERTa-XSmall}. 

\begin{table}[htbp]
\caption{Performance Metrics of DeBERTa-XSmall}
\begin{center}
\resizebox{\columnwidth}{!}{%
\begin{tabular}{|c|c|c|c|}
\hline
\textbf{Fold}&
\textbf{Training Loss}& 
\textbf{Validation Loss}&
\textbf{Pearson Correlation}\\ \hline
1& 
0.039200 & 
0.030078 &
0.774637 \\ 
\hline

2& 
0.039200 & 
0.031391 &
0.765988 \\ 
\hline

3& 
0.038800 & 
0.029105 &
0.780142 \\ 
\hline

4& 
0.038700 & 
0.031934 &
0.755139 \\ 
\hline

\end{tabular}
}
\end{center}
\label{DeBERTa-XSmall}
\end{table}

\section{Conclusion}
This paper experimented with traditional and transformer-based approaches for semantic similarity modeling on large corpora. We also compared our methodology with existing techniques, and the results demonstrated the improved performance of the model. The proposed methodology also illustrated context extraction and showed its importance in similarity modeling. In the following aspects, the execution time and memory could be optimized, thus leading to enhanced training. Also, the architecture of the existing model could be improved, thus leading to enhanced performance.

\bibliographystyle{IEEEtran}
\bibliography{IAC}

\end{document}